# Research on Fitness Function of Two Evolution Algorithms Using for Neutron Spectrum Unfolding

**Rui Li**
*Chengdu University of Technology, Chengdu 610059, China*

**Jianbo Yang\*, Xianguo Tuo and Rui Shi**
*Chengdu University of Technology, Chengdu 610059, China*
*Sichuan University of Science and Engineering, Zigong 643000, China*
*Applied Nuclear Technology in Geosciences Key Laboratory of Sichuan Province，Chengdu 610059, China*

Using evolution algorithms to unfold the neutron energy spectrum, fitness function design is an important fundamental work for evaluating the quality of the solution, but it has not attracted much attention. In this work, we investigated the performance of 8 fitness functions attached to genetic algorithm (GA) and differential evolution algorithm (DEA) used for unfolding four neutron spectra selected from IAEA 403 report. Experiments show that the fitness functions with a maximum in GA can limit the ability of the population to percept the fitness change but the ability can be made up in DEA, and the fitness function with a feature penalty term help to improve the performance of solutions, and the fitness function using the standard deviation and the Chi-Squared shows the balance between algorithm and spectra. The results also show that the DEA has good potential for neutron energy spectrum unfolding. The purpose of this work is to provide evidence for structuring and modifying the fitness functions, and some genetic operations that should be paid attention were suggested for using the fitness function to unfold neutron spectra.

# I. INTRODUCTION

Bonner multi-sphere neutron energy spectrometer (BSS) was introduced by Bramblett et al. [1], which consists of multiple moderator spheres with different thicknesses and a thermal neutron detector [2,3]. However, the neutron spectrum cannot be read from the detector directly but need the neutron spectrum unfolding method to calculate, and the unfolding process is expressed by the first kind Fredholm integral equation, which could be represented in a discrete form as Eq. (1) [4].

$$C_j = \sum_{i=1}^{n} R_{ij}\varphi_i + \varepsilon_i \quad j = 1,2,3,...m \tag{1}$$

Where $C_j$ is neutron count reading from the *j*th detection unit, and $R_{ij}$ is the response of *j*th detection unit to neutrons of *i*th energy group simulated from Monte Carol code generally, and $\varphi_i$ is the neutron fluence of *i*th energy group, and $\varepsilon_i$ is the detect error of *j*th detection unit. $C_j$ and $R_{ij}$ will be the input of the unfolding algorithm, and $\varphi_i$ be the output. Normally, the number of detect units is far smaller than the number of energy groups (i.e., $m \ll n$, in this work, m=15, n=53).

To obtain an approximate neutron spectrum, some artificial intelligence optimization methods can be used, such as neural networks method [5], particle swarm optimization [6, 7], genetic algorithm [8], and differential evolution algorithm in this work. The quality of the solution and the algorithm performance is often strongly related to fitness function in GA and DEA [9]. In the current literature, however, few detailed studies have been carried out on the fitness function for neutron unfolding problem. In this work, the structural characteristics of different fitness functions and their effects on the quality of the solution were investigated in GA and DEA frames.

# II. MATERIAL AND METHODS

## 1. Genetic algorithm

GA [10] imitates the natural evolution of biology which has been introduced by Darwin (i.e., "survival of the fittest"). Unlike the common linear optimization technique using for neuron spectrum unfolding, the GA optimization is capable of finding the global optima in a large multi-dimensional solution space [13].

**Mutation**, in this work, a single point mutation was performed to introduce new genes into the individual. Firstly, randomly select an individual from the current population following the mutation probability, then, randomly select a locus from the selected individual to mutate to a real number $\in (0, \min(C_j/R_{ij}))$.

**Crossover** is a way for the genes, including excellent genes, of parents can be passed onto the offspring. The whole arithmetic crossover operator was performed [11]

$$\begin{aligned} o^1 &= u \cdot p^1 + (1-u) \cdot p^2 \\ o^2 &= (1-u) \cdot p^1 + u \cdot p^2 \end{aligned} \quad (2)$$

where $o^1$ and $o^2$ are the offspring, $p^1$ and $p^2$ are the parents selected randomly from the population, $u$ is a rand number $\in (0,1)$.

**Selection** is the elimination mechanism in GA. Russian roulette method was used in this work, which results in individuals with higher fitness have more chance to be carried forward to the next generation. Before the selection, individuals were evaluated by the fitness function.

## 2. Differential evolution algorithm

DEA was first proposed by Storn et al. [12], also consists of mutation, crossover, and selection, but different from specific evolution strategy.

**Mutation** in DEA does not send the offspring into the population immediately but will be temporarily stored for transferring to the crossover operation. Three individuals were randomly selected from the population, and the mutation was performed as the following Eq. (3) [12]

$$\mathbf{x}^{tem} = \mathbf{x}^{i1} + F \cdot (\mathbf{x}^{i2} + \mathbf{x}^{i3}) \quad (3)$$

where $\mathbf{x}^{tem}$ is the temporary individual, $F$ is a scale factor $\in (0,2)$, in this work, $F$=0.5, $\mathbf{x}^{i1}$, $\mathbf{x}^{i2}$ and $\mathbf{x}^{i3}$ are selected individuals from the population, and $i1 \neq i2 \neq i3$.

**Crossover** operation is carried out based on mutation. A target individual $\mathbf{x}^{tar}$ and the temporary individual $\mathbf{x}^{tem}$ cross by sending the genes to offspring [12]:

$$x_i^o = \begin{cases} x_i^{tem}, & u < \text{pc} \\ x_i^{tar}, & \text{otherwise} \end{cases} \quad (4)$$

where $\mathbf{x}^o = [x_1^o, x_2^o, x_3^o, ... x_n^o]^T$ is the offspring will be sent to the selection, $u$ is a random number $\in (0,1)$, pc is the crossover probability.

**Selection** of DEA carries strictly the individual forward to the next generation with higher fitness between the target individual $\mathbf{x}^{tar}$ and the offspring $\mathbf{x}^o$, before the selection, both of them were evaluated by the fitness function.

From the above, both in GA and DEA, selection decides which individual can survive to the next generation based on the fitness. Therefore, the fitness function controls the evolution of the population.

## 3. Fitness function

In the application of neutron spectrum unfolding, the fitness function is used to estimate how close between the solution and the reference neutron spectrum. In this work, we found that the evolution behavior and the quality of solution varied significantly from different fitness functions, and the same fitness function in the DEA and GA also shows the makeable differences, the fitness functions are given in Table 1.

Table 1. fitness function.

| Fun No. | Fitness Function |
|---|---|
| F1 [13] | $f_1 = \sum_{j=1}^{m}\left[\beta_1 - \left(T_j \big/ \sum_{j=1}^{m} R_{ij}\varphi_i^{cal} + C_j\right)^2\right]$ |
| F2 [14, 15] | $f_2 = 1 \big/ \sum_{j=1}^{m} T_j^2 / C_j^2$ |
| F3 [14] | $f_3 = \max(\sum_{j=1}^{m} T_j^2/C_j^2) \cdot 2 - \sum_{j=1}^{m} T_j^2/C_j^2$ |
| F4 [16] | $f_4 = 1 \big/ (\sum_{j=1}^{m} T_j^2 + \beta_4 \cdot p_1)$ |
| F5 | $f_5 = 1 \big/ \delta((\sum_{j=1}^{m} R_{ij}\varphi_i^{cal}) \big/ C_j)$ |
| F6 [4] | $f_6 = \beta_6 + \sum_{j=1}^{m} -(T_j/C_j)^2$ |
| F7 [17] | $f_7 = 1 \big/ \sqrt{\sum_{j=1}^{m} C_j^2 \big/ \sum_{j=1}^{m} T_j^2}$ |
| F8 | $f_8 = 1 \big/ (\sum_{j=1}^{m}(T_j/C_j)^2 + 0.5 \cdot \beta_8 \cdot (p_1 + p_2))$ |

Where $T_j = (\sum_{j=1}^{m} R_{ij}\varphi_i^{cal}) - C_j$, and $\varphi_i^{cal}$ is the neutron fluence of *i*th energy group calculated from unfolding algorithm, and $\beta_1 = 0.1 \cdot \beta_4 = \beta_6 = 100$, $p_1 = \sum_{i=2}^{n}(\varphi_i^{cal} - \varphi_{i-1}^{cal})^2$, $p_2 = \sum_{i=2}^{n-1}(\varphi_{i-1}^{cal} - 2 \cdot \varphi_i^{cal} + \varphi_{i+1}^{cal})^2$, $p_1$ and $p_2$ are used to calculate the continuity of energy spectrum [8, 16]. F5 and F8 are varieties from literature [8, 18].

In GA, 500 individuals were initialized randomly, and the maximum iteration was 5000, and the mutation probability was 0.1, and the crossover probability was 0.9. The parameters in the DEA were set the same as in GA but without mutation probability.

## 4. Reference spectra and Quality factors

It was mentioned in the literature [4, 16, 18] that neutron energy spectra usually have good continuity. Therefore, to make the results more representativeness in this work, as shown in Fig. 1, we selected four reference neutron spectra with different shape and continuity, and response matrix reported by IAEA 403 [18]. The continuity of the spectra was calculated as $\sum_{i=2}^{n}(\varphi_i^{cal} - \varphi_{i-1}^{cal})^2$, and the continuity of the spectra is 1.27, 0.1, 0.01, and 0.001 from Spec1 to Spec4 becoming better. In the interval of $10^{-9} \sim 15.8\text{MeV}$, 53 energy groups are divided, and 15 detection units are used. The result of the convolution of reference spectra and response matrix simulating neuron counts reading from detection units, and the random noise with a standard deviation of 5% was intentionally added.

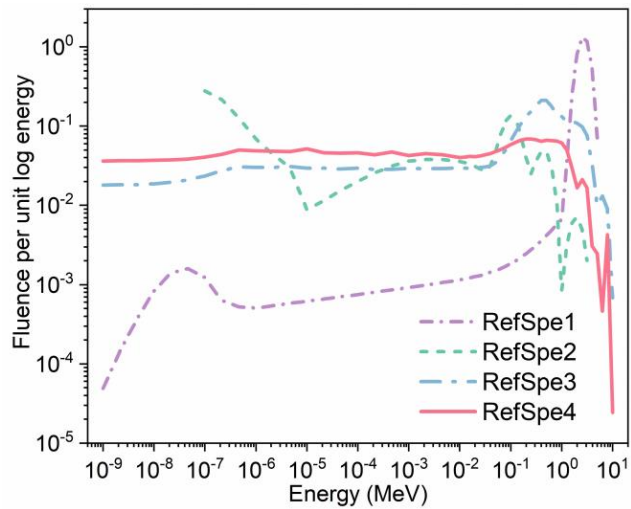

Fig. 1. Four reference spectra

Spectrum quality factor (Qs) [4] was used to evaluate the shape difference between the solution and the reference energy spectrum (true value) in this work.

$$Qs = 100 \cdot \sqrt{\frac{\sum_{i=1}^{n}\left(\varphi_i^{ref} - \varphi_i^{cal}\right)^2}{\sum_{i=1}^{n}(\varphi_i^{cal})^2}} \qquad (5)$$

Where $\varphi_i^{ref}$ is the fluence of the *i*th energy group of the reference energy spectrum. A perfect solution produces Qs = 0.

## III. RESULTS AND DISCUSSION

Static distribution is shown in Fig. 2, Taking the fluence of each energy group of Spec4 as the center, random and uniform sampling was performed within $(0, \min(C_j/R_{ij}))$. Each experiment of fitness function sampled 200 thousand energy spectra to build a static population, simulating different quality solutions during the evolution process.

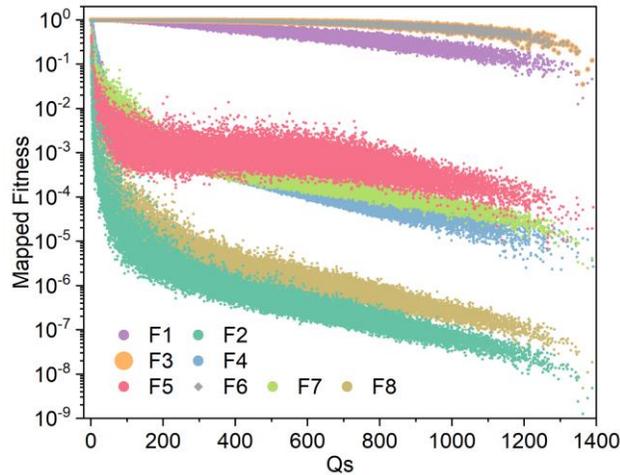

Fig. 2. Qs-fitness distribution of a static population, for illustration, all the fitness values from different functions were mapped into the (0, 1).

The individual with highest fitness produced the best Qs means that all the fitness functions were appropriate under ideal evolution, and the distribution of fitness functions varies from the form of functions. Specifically, F5 is different from other functions with a bulge (Qs≈600). The span of fitness function values of F2, F8, F4, and F7 increases successively, Qs is about close to 0, the distribution is about steep. In other words, in a collection of individuals close to the true value (Qs≈0), a slight change in the shape of the spectrum can cause a drastic change in the value of this fitness functions. Conversely, the distribution of F1, F3, and F6 are flatter, because there is a constant term in each function that limits the maximum value of the fitness. The fitness value of group 2 (F4, F7, F8, and F2) and group 3 (F1, F3, and F6) can be described as G3≈Max-G2.

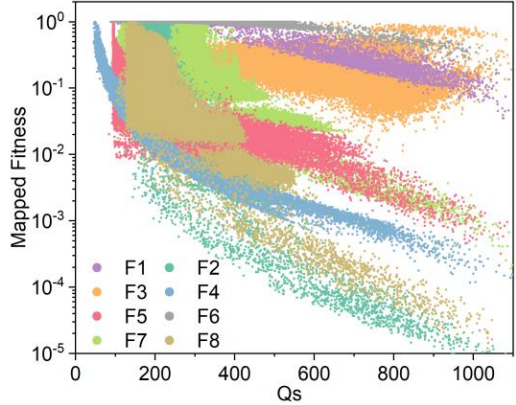

Fig. 3. Qs-fitness distribution of dynamic population in GA using Spec4, during evolving, the fitness was ranked after each iteration, and 10% of individuals in the current population are uniformly extracted from the best to the worst.

For each fitness function in Fig. 3, the same fitness level has a certain range of the Qs, that is to say, one value of fitness map to several solutions with different spectrum shapes. In the same way, individuals with the same Qs can produce a wide range of fitness. In this case, selection operation cannot distinguish the difference between the better individuals and the inferior individuals based on fitness. Moreover, inferior individuals with high fitness can deceive the algorithm, therefore, the optimal individual (Qs closer to 0) in the F2, F7, F8, and F5 are on the right side of the highest fitness.

There is a maximum in F1 and F6, in the mid-late evolution, the individuals are close, the opportunity between the better individuals and others to survive is close. In other words, the algorithm cannot accurately identify and keep high-quality individuals, resulting in an optimal individual with poor quality after 5000 iterations.

F5 is the most different from other functions because F5 takes the standard error as the fitness, and more attention should be paid. As we all know, the standard deviation is used to measure the degree of dispersion of the data, and two data samples with a larger difference in mean can also get a smaller standard deviation. In the process of neutron energy spectrum unfolding, if the neutron counts of detection units calculated by the solution are all very small or very large, according to F5, a smaller standard deviation, can also be obtained. However, neither of these two situations is what we expect. Therefore, after a lot of experiments, we suggest setting a maximum to limit the fitness of the solution, such as 10000 in this work, or limiting the error of each count ($(\sum_{j=1}^{m} R_{ij}\varphi_i^{cal})/C_j$) to be less than a certain value, such as 0.0001.

The first item ($\max(\sum_{j=1}^{m} T_j^2/C_j^2)\cdot 2$) within F3 depended on the individual closest to the reference spectrum in ever single generation, called elite individual. If the elite individual changes between two neighboring generations, the fitness of the remaining individuals will also change, even if these individuals themselves have not changed. Therefore, the fitness between t generation and t+1 generation may not comparable, and the algorithm outputs the optimal solution of the last iteration instead of history optimal.

For a fitness function, we also hope that the fitness function can show the same excellent performance for different spectra. The statistical results of Qs of history optimal individuals are shown in Fig. 4, and each result was coming from 20 independent runs.

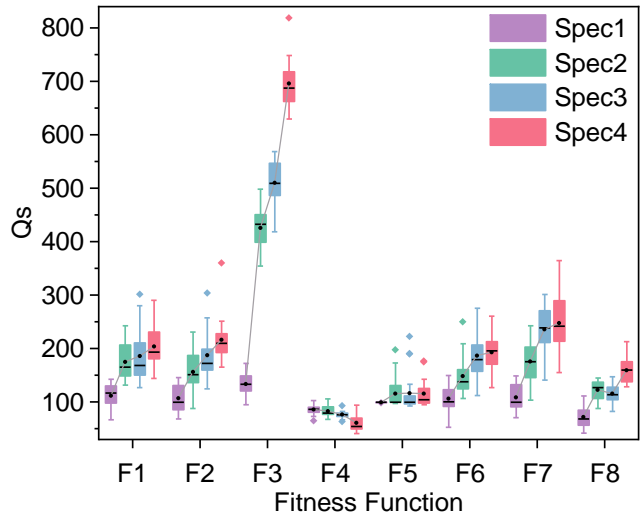

Fig. 4. Qs of history optimal individuals in GA

Most neutron energy spectra have good continuity, unless in some extreme cases [4, 16, 18]. As the continuity of the reference spectrum becomes better from Spec1 to Spec4, Qs of each solution becomes better in F4. Because F4 maintains a great continuity penalty factor to eliminate individuals with poor continuity in time. The ability of F2, F5, F7, and F8 to perceive the shape shake are weaker than F1, F3, and F6, but better results can be obtained.

To investigate the application of the fitness function in different algorithm frameworks, this work proposed to use the DEA to unfold the neutron energy spectrum. It can be considered as the feasibility verification of DEA for neutron spectrum unfolding.

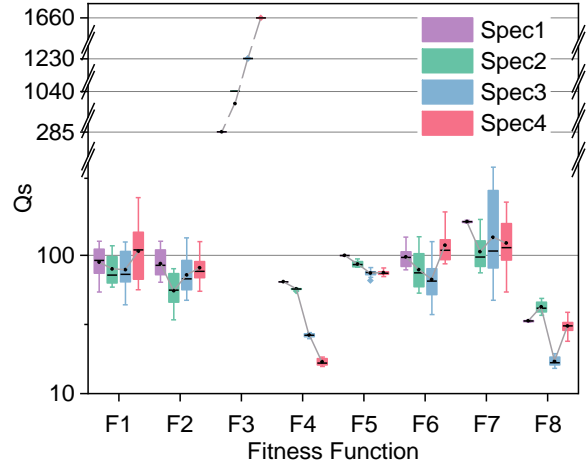

Fig. 5. Qs of history optimal individuals in DEA

As shown in Fig. 5, each result was also coming from 20 independent runs. Except F3, the quality and the robustness of results has been improved in the DEA, and there is a surprise that F1 and F6 have the greatest performance improvement in DEA compared with the GA, because in DEA, only the individual with higher fitness will be remain, which eliminated the impact of inferior individuals on selection operation. All those improvements in DEA may provide some ideas for modifying GA's genetic operation strategy, that eliminates inferior individuals in time or increase the proportion of better individuals.

As shown in Table 2 and Fig. 7, the Qs of the best solution of each fitness function in GA and DEA is given. The best solutions of the DEA are better than GA in the case of reference Spec2, Spec3, and Spec4, as for Spec1, the Qs of the best solutions in DEA and GA are similar.

F4 and F8 adopted a characteristic penalty term on the continuity of the energy spectrum into the fitness function, and output the best results in different energy spectrum expansions. As the continuity of the reference energy spectrum becomes better, it means that the F4 and F8 are easier to distinguish the characteristics of the energy spectrum, resulting in better energy spectrum unfolding performance.

Table 2. Qs of the best solution of the reference neutron spectrum of each fitness function in GA and DEA.

| | GA | | | | DEA | | | |
|---|---|---|---|---|---|---|---|---|
| | Spec1 | Spec2 | Spec3 | Spec4 | Spec1 | Spec2 | Spec3 | Spec4 |
| F1 | 67.58 | 132.0 | 127.4 | 145.1 | 76.21 | 79.47 | 67.95 | 77.79 |
| F2 | 69.27 | 88.80 | 125.4 | 166.0 | 82.57 | **58.16** | 70.87 | 76.82 |
| F3 | 95.25 | 354.9 | 418.8 | 629.9 | 285.2 | 844.5 | 1230 | 1659 |
| F4 | 65.37 | **68.53** | **63.59** | **41.98** | 82.87 | 76.74 | 45.90 | **27.90** |
| F5 | 98.63 | 97.98 | 92.88 | 95.55 | 99.80 | 92.26 | 83.66 | 86.17 |
| F6 | **53.47** | 107.5 | 112.6 | 127.6 | 90.64 | 75.46 | 61.64 | 94.55 |
| F7 | 71.21 | 103.8 | 141.7 | 155.4 | 120.9 | 88.74 | 70.92 | 76.12 |
| F8 | 42.80 | 88.37 | 83.22 | 129.4 | **56.57** | 61.05 | **26.61** | 44.32 |

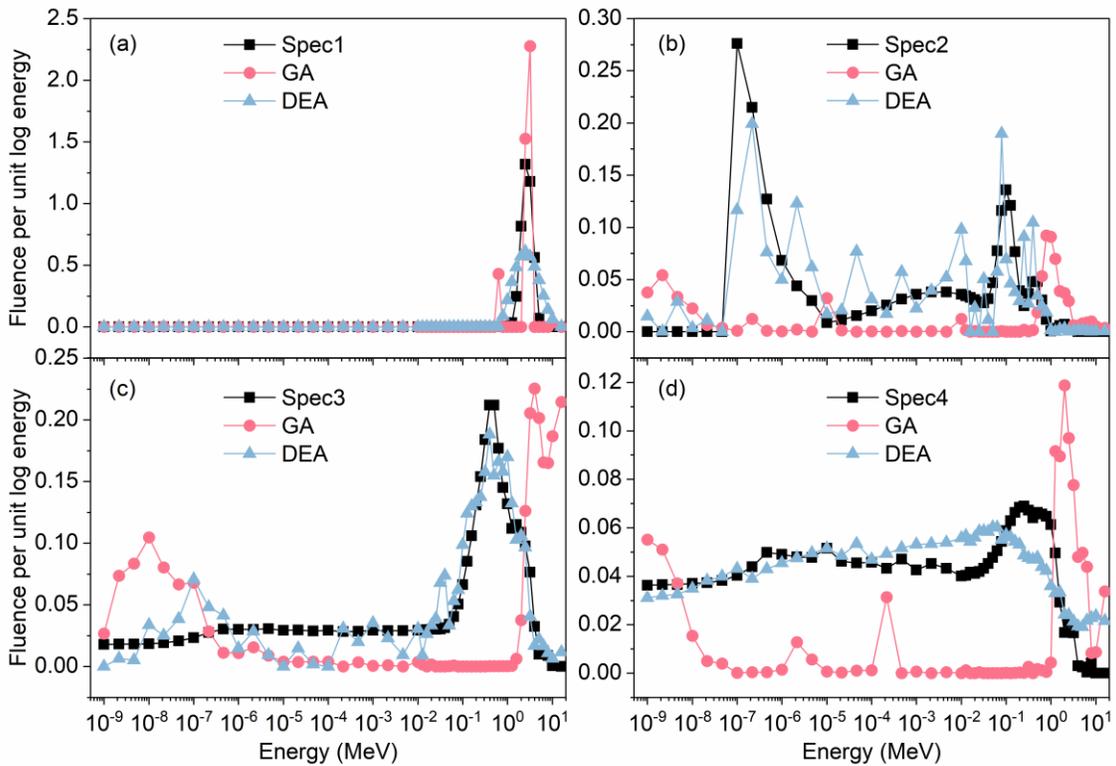

Fig. 7. Unfolded spectra, the unfolded spectra with the lowest Qs value in Table 2, (a) is the result of the F6 in GA and the F8 in DEA, (b) is the result of the F4 in GA and the F2 in DEA, (c) is the result of the F4 in GA and the F8 in DEA, and (d) is the result of the F4 in GA and the F4 in DEA.

As shown in Fig. 6, the Qs of the best solution fluctuated with iteration greatly, especially in Fig. 6(b), and the Qs of the final optimal solution was almost not the lowest value. Calculations after the lowest Qs are invalid or even harmful. It is worth noting that the quality change of the energy spectrum is closely related to the shape change of the energy spectrum. Therefore, "shape"-related constraints are suggested to adopt into the iterative termination and restart criteria during the unfolding.

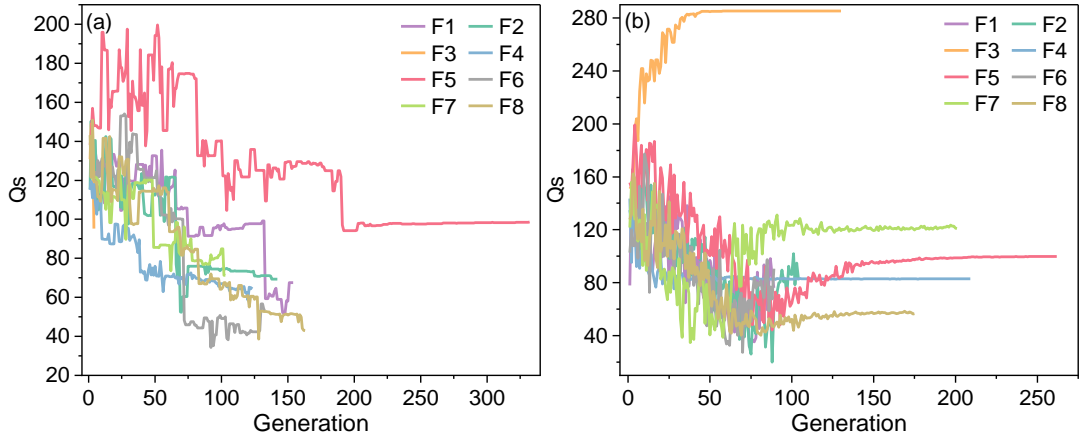

Fig. 6. Qs of the best solution of the Spec1 in GA and DEA of each the effective evolutionary generation (the fitness of the best solution of the current generation is bigger than the previous generation), (a) is the result of the GA, and (b) is the result of the DEA.

## IV. CONCLUSION

This work investigated the evolution behavior of the eight different fitness functions used in the GA and the DEA, and four different spectra in IAEA 403 report were selected to perform, and three conclusions can be obtained:

1. The fitness functions are designed for GA can also be used in DEA, and confirmed that the good potential of DEA for neutron energy spectrum unfolding.

2. Specifically, the fitness functions such as F1 and F6 with an upper are not friendly to GA, which will weaken the ability of perception and response to fitness change, therefore, we recommend that pay attention to increase the selection pressure of GA when using; F2, and F7 may need other genetic techniques to make up; For F3, A dynamic maximum makes the function to output the optimal solution of the last iteration instead of history optimal, so, there is not enough courage to use in most cases; For F4 and F8, functions with a continuity penalty term show the best solution, which means the fitness function designed for the characteristics of the spectrum can improve the performance. Furthermore, F4 is suitable for GA, while F8 is suitable for DEA; For F5, since the standard error of the neutron counting error is used as the fitness, it is necessary to cooperate with more constraints, such as limiting the maximum fitness of the best solution or limiting the error of each detect unit, and shows that a good balance between the algorithm frameworks and the energy spectra.

3. The fitness increase, while the quality of the energy spectrum may not improve, which results in a waste of computing time. "shape change" as the condition for stopping iteration or restart is recommended, it is significant for applications of online and fast unfolding.

## ACKNOWLEDGMENTS

This work was supported by the National Natural Science Foundation of China (No.41774120) and Sichuan Science and Technology Program (2018TJPT0008, 19ZDYF1561) and Opening Fund of Provincial Key Lab of Applied Nuclear Techniques in Geosciences (No. gnzds201903) of Applied Nuclear Technology in Geosciences Key Laboratory of Sichuan Province, Chengdu University of Technology

## REFERENCE

[1] R. L. Bramblett, R. I. Ewing, and T. W. Bonner, Nuclear Instruments and Methods. **9**, 1 (1960).
[2] J. B. Yang *et al.*, Patent No. ZL201610264803.7 (2018) (in Chinese).
[3] J. B. Yang *et al.*, Patent No. US10656292B2 (2018).
[4] D. W. Freeman, D. R. Edwards, and A. E. Bolon, Nucl. Instrum. Methods Phys. Res. Sect. A-Accel. Spectrom. Dect. Assoc. Equip. **425**, 549 (1999).
[5] K. Chang *et al.*, J. Korean Phys. Soc. **74**, 542 (2019).
[6] H. Shahabinejad, and M. Sohrabpour, Radiat. Phys. Chem. **136**, 9 (2017).
[7] D. Zhao *et al.*, Nucl. Instrum. Methods Phys. Res. Sect. A-Accel. Spectrom. Dect. Assoc. Equip. **933**, 56 (2019).
[8] J. Wang *et al.*, Appl. Radiat. Isot. **147**, 136 (2019).
[9] S. Kazarlis, V. Petridis, in *Parallel Problem Solving from Nature-PPSN V*, ed. by Eiben A.E., Bäck T., Schoenauer M., Schwefel HP. (Netherlands, September 27-30, 1998), Vol. 1498, p. 211.
[10] D. E. Goldberg, *Genetic Algorithms in Search, Optimization, and Machine Learning*, (Addison-Wesley Publishing Company, Boton, 1989), Vol, 1, Chap, 1, pp.1-15.
[11] A.E. Eiben, and J.E. Smith, *Introduction to Evolutionary Computing*, (Springer-Verlag Berlin Heidelberg, 2015). Vol, 4, Chap, 4, pp. 66.
[12] R. Storn, and K. Price, J. Glob. Optim. **11**, 341 (1997).
[13] J. A. Santos *et al.*, Appl. Radiat. Isot. **71**, 81 (2012).
[14] X. Wang *et al.*, Nucl. Sci. Tech. **25**, 1 (2014).
[15] S. M. T. Hoang *et al.*, J. Radioanal. Nucl. Chem. **318**, 631 (2018).
[16] D. WANG, B. HE, and Q. H. ZHANG, Atomic Energy Science and Technology. **44**, 1270 (2010) (in Chinese).

[17] H. Shahabinejad, S. A. Hosseini, and M. Sohrabpour, Nucl. Instrum. Methods Phys. Res. Sect. A-Accel. Spectrom. Dect. Assoc. Equip. **811**, 82 (2016).
[18] IAEA, Compendium of Neutron Spectra and Detector Responses for Radiation Protection Purposes, No. 403, 2001.